%% file: main.tex
\documentclass{article} 
\usepackage{colm2024_conference}

\usepackage{wrapfig}
\usepackage{multirow}
\usepackage{microtype}
\usepackage{hyperref}
\usepackage{url}
\usepackage{booktabs}
\usepackage{tikz}

\usepackage{subfigure}
\usepackage{caption}

\usepackage{changes}
\usepackage{graphicx}
\usepackage[frozencache,cachedir=.]{minted}
\usepackage{paralist}
\usepackage{xspace}
\usepackage{amsthm}
\usepackage{amsmath, amssymb}  
\usepackage{enumitem} 
\usepackage[export]{adjustbox} 
\usepackage{listings}
\input{math_commands}

\newcommand{\gpt}{\textsc{GPT}\xspace}
\newcommand{\gptf}{\textsc{GPT-4}\xspace}
\newcommand{\gptt}{\textsc{GPT-3.5}\xspace}
\newcommand{\llama}{\textsc{LlaMA}\xspace}
\newcommand{\llamas}{\textsc{LlaMA-7B}\xspace}
\newcommand{\llamat}{\textsc{LlaMA-13B}\xspace}
\newcommand{\llava}{\textsc{LlaVA}\xspace}
\newcommand{\gemma}{\textsc{Gemma-7B}\xspace}
\newcommand{\llmtime}{\textsc{LLM-Time}\xspace}

\theoremstyle{definition}
\newtheorem{definition}{Definition}[section]

\newcommand{\ie}[0]{{\em i.e.,~}}

\newcommand{\qt}[1]{``#1''}

\usepackage{titlesec}
\titlespacing*{\section}{0pt}{0.3\baselineskip}{0.2\baselineskip}
\titlespacing*{\subsection}{0pt}{0.3\baselineskip}{0.2\baselineskip}

\newcommand{\xhdr}[1]{\vspace{1.7mm}\noindent{{\bf #1.}}}

\title{Language Models Still Struggle to Zero-shot Reason about Time Series}

\author{Mike A. Merrill\\
University of Washington \\
\texttt{mikeam@cs.washington.edu} \\
\And
Mingtian Tan\\
University of Virginia\\
\texttt{wtd3gz@virginia.edu}\\
\And
Vinayak Gupta \\
University of Washington\\
\texttt{vinayak@cs.washington.edu}\\
\And
Tom Hartvigsen \\
University of Virginia\\
\texttt{hartvigsen@virginia.edu}
\And
Tim Althoff\\
University of Washington\\
\texttt{althoff@cs.washington}
}

\colmfinalcopy 

\begin{document}
\maketitle

\begin{abstract}
Time series are critical for decision-making in fields like finance and healthcare.
Their importance has driven a recent influx of works passing time series into language models, leading to non-trivial forecasting on some datasets.
But it remains unknown whether non-trivial forecasting implies that language models can reason about time series.
To address this gap, we generate a first-of-its-kind evaluation framework for time series reasoning, including formal tasks and a corresponding dataset of multi-scale time series paired with text captions across ten domains.
Using these data, we probe whether language models achieve three forms of reasoning:
(1) \textit{Etiological Reasoning}---given an input time series, can the language model identify the scenario that most likely created it?
(2) \textit{Question Answering}---can a language model answer factual questions about time series?
(3) \textit{Context-Aided Forecasting}---does highly relevant textual context improve a language model's time series forecasts?
We find that otherwise highly-capable language models demonstrate surprisingly limited time series reasoning: they score marginally above random on etiological and question answering tasks (up to 30 percentage points worse than humans) and show modest success in using context to improve forecasting. These weakness showcase that time series reasoning is an impactful, yet deeply underdeveloped direction for language model research. We also make our datasets and code public at to support further research in this direction at \url{https://github.com/behavioral-data/TSandLanguage}.
\end{abstract}

\input{00-figs}
\input{001_introduction}

\input{003_definitions}
\input{004_dataset}

\input{005_experiments}

\input{002_related_work}

\input{006_conclusion}
\subsubsection*{Acknowledgments}
This research was supported in part by NSF CAREER IIS-2142794, Bill \& Melinda Gates Foundation (INV-004841), NSF IIS-1901386, NSF CNS-2025022, the Microsoft Accelerating Foundation Models Research Program, and UW eScience Azure Cloud Computing support.

\clearpage
\bibliographystyle{colm2024_conference}
\bibliography{refs}

\clearpage

\setcounter{table}{0}
\setcounter{figure}{0}
\renewcommand{\thetable}{A.\arabic{table}}
\renewcommand{\thefigure}{A.\arabic{figure}}
\appendix

\input{100-appendix}

\end{document}

%% file: math_commands.tex
\usepackage{amsmath,amsfonts,bm}

\def\eqref#1{equation~\ref{#1}}

\def\1{\bm{1}}

\def\sR{{\mathbb{R}}}



%% file: 00-figs.tex
\newcommand{\figSummResults}{
    \begin{figure*}
        \includegraphics[width=\linewidth]{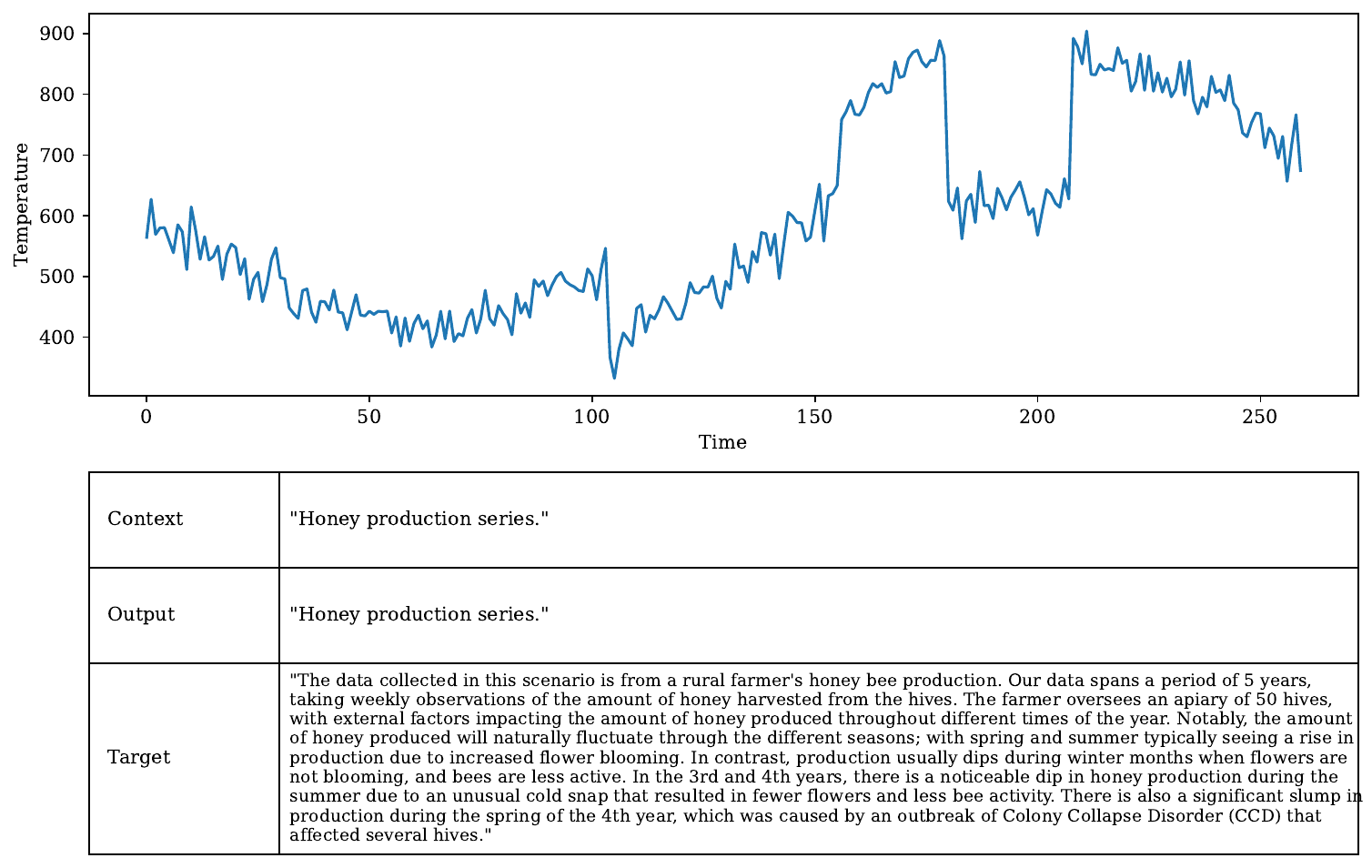}
        \caption{Simple summarization results.}
        \label{fig:summResults}
    \end{figure*}
}

\newcommand{\figGenCode}{
    \begin{minted}{python}
        
        test
    \end{minted}
}

\newcommand{\figOneOfEach}{
\begin{figure*}
    \centering
    \includegraphics[width=\linewidth]{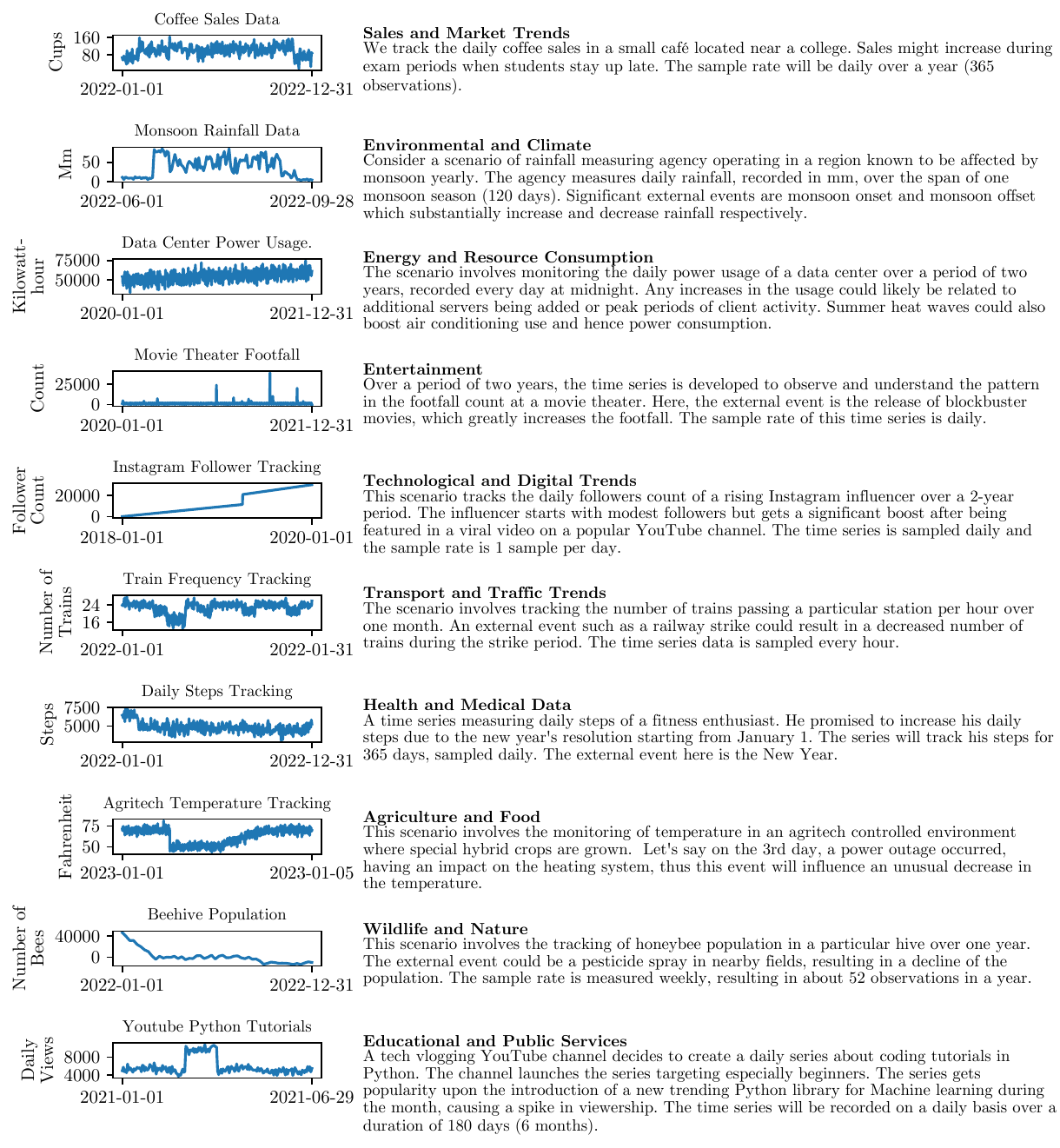}
    \caption{One scenario from each of our ten categories (Section \ref{sec:dataset}). \vspace{-0.5cm}}
    \label{fig:OneOfEach}
\end{figure*}
}

\newcommand{\figreal}{
\begin{figure*}[t!]
    \centering
    \includegraphics[width=0.6\linewidth]{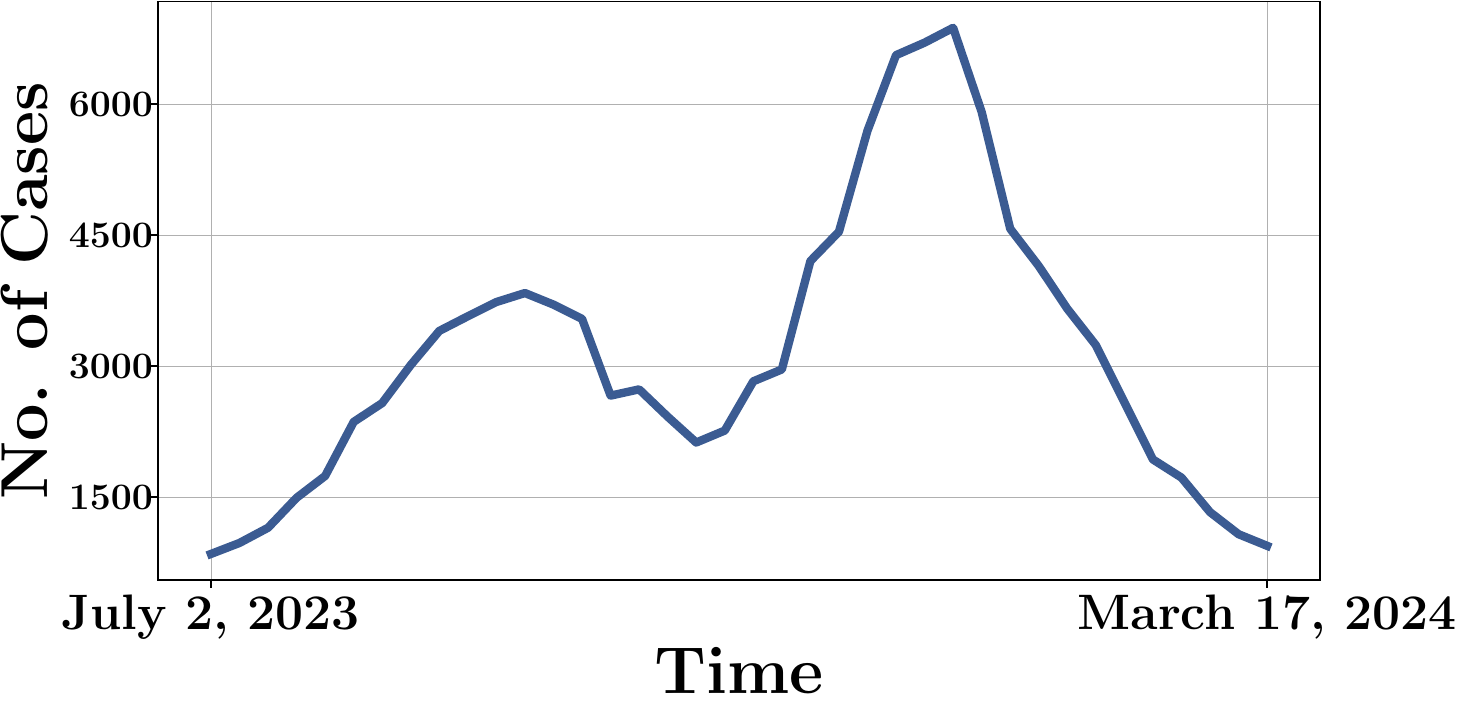}
    \caption{COVID-19 cases for Massachusetts: An example of real data time series used for generating MCQs.}
    \label{fig:covid}
\end{figure*}
}

\newcommand{\figsinglets}{
\begin{figure*}[t!]
    \centering
    \includegraphics[width=0.6\linewidth]{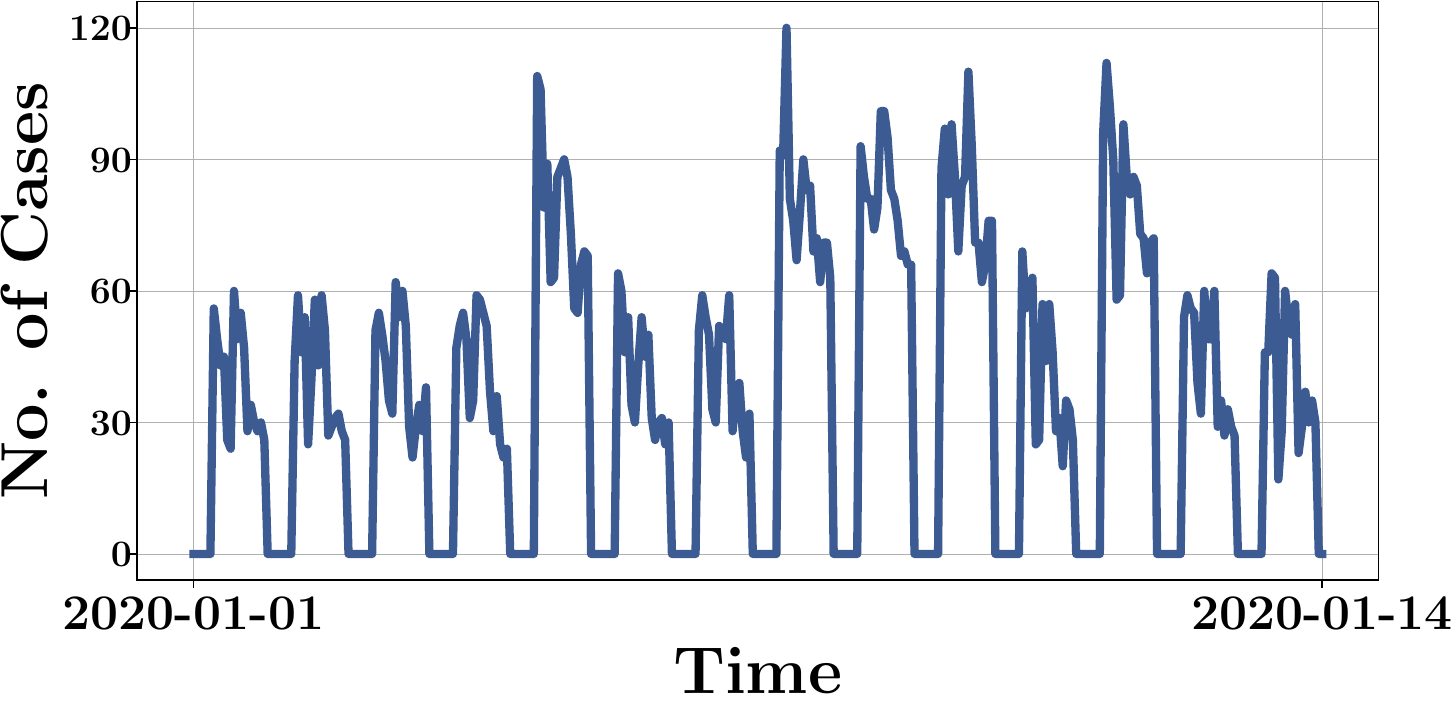}
    \caption{An example time-series with description: 'Customer counts at a cafe following a TV feature over two week period.'}
    \label{fig:singlets}
\end{figure*}
}

\newcommand{\figScenario}{
    \begin{figure*}
    \centering
        \includegraphics[width=\linewidth]{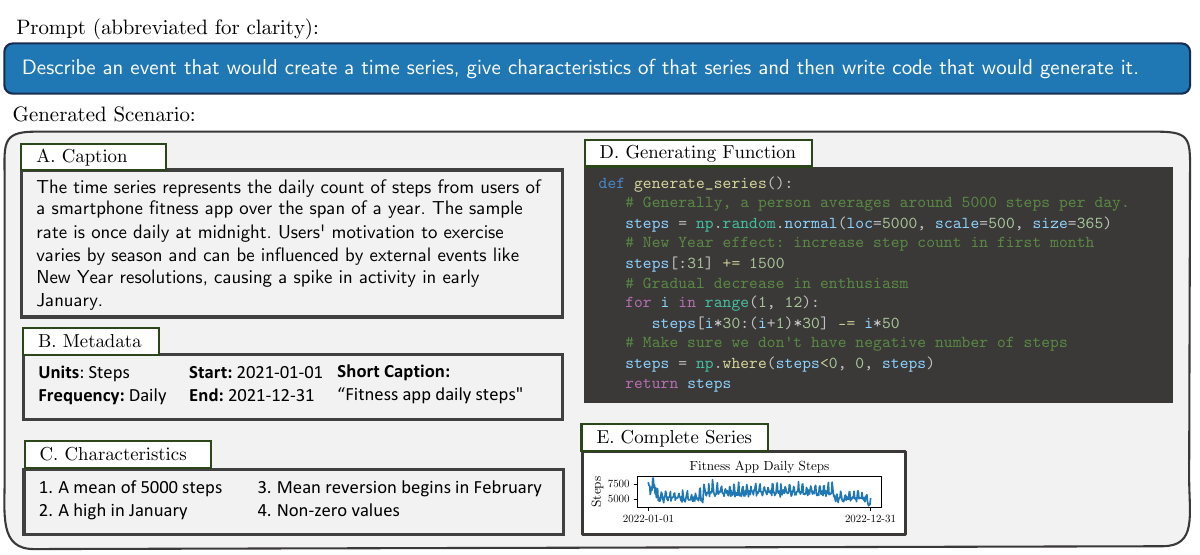}
        \caption{We generate realistic time series and text pairs by querying GPT-4 for code that can be executed to generate the signal (Section \ref{sec:dataset}).}
        \label{fig:summResults}
    \end{figure*}
}

\newcommand{\figDonut}{
\begin{figure}
    \centering
    \includegraphics[width=\linewidth]{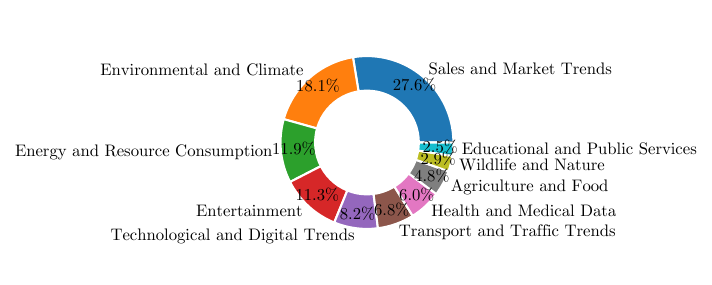}
    \caption{Portion of scenario categories in our generated dataset (Section \ref{sec:dataset}).}
    \label{fig:Donut}
\end{figure}
}

\newcommand{\figTextVSCode}{
\begin{figure*}
    \centering
    \includegraphics[width=\linewidth]{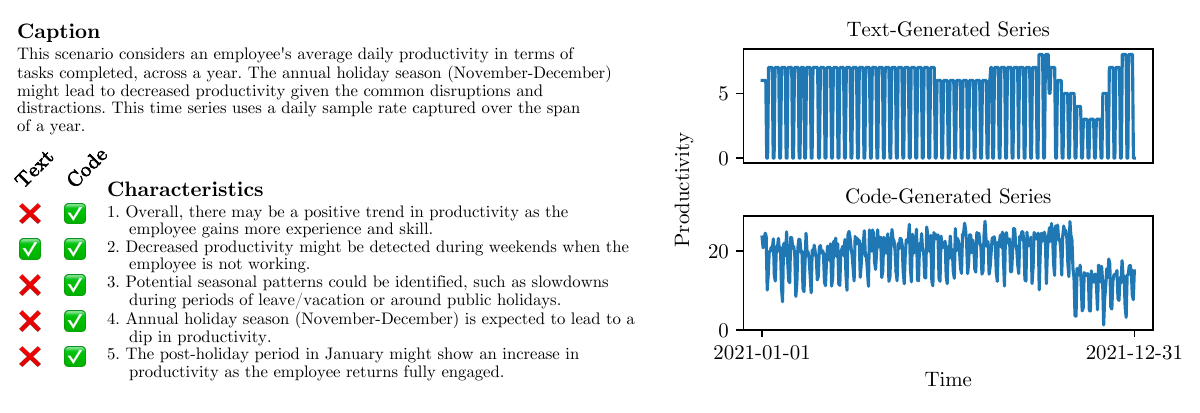}
    \caption{By simulating time series with GPT-4-generated code (rather than generating the series directly from the model itself) we're able to produce substantially more complex data which better represent realistic scenarios.}
    \label{fig:TextCSCode}
\end{figure*}
}

\newcommand{\figHero}{
\begin{figure*}
   \centering
    \includegraphics[width=0.9\linewidth]{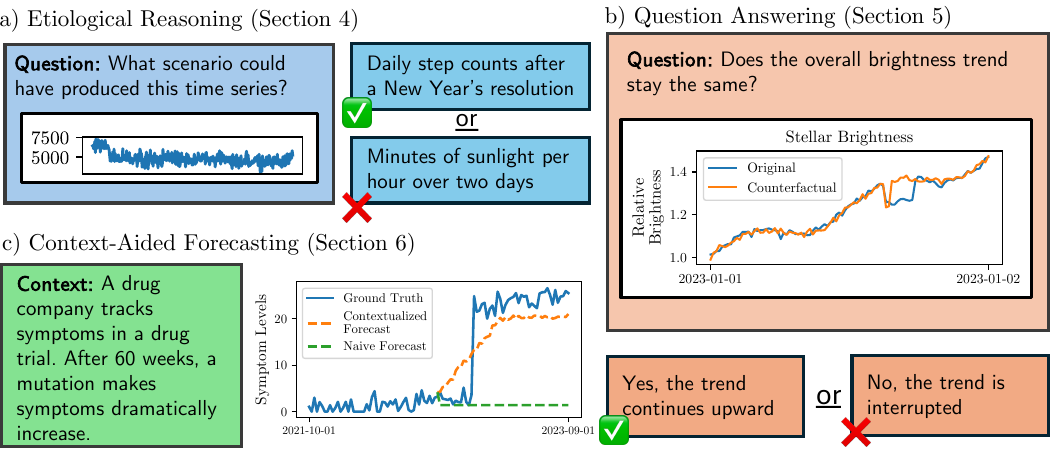}
    \caption{The three forms of time series reasoning (Section \ref{sec:forms_of_reasoning}). \vspace{-0.5cm}}
    \label{fig:hero}
\end{figure*}
}

\newcommand{\figAnnotatorTool}{
\begin{figure*}
    \centering
    \includegraphics[width=\linewidth]{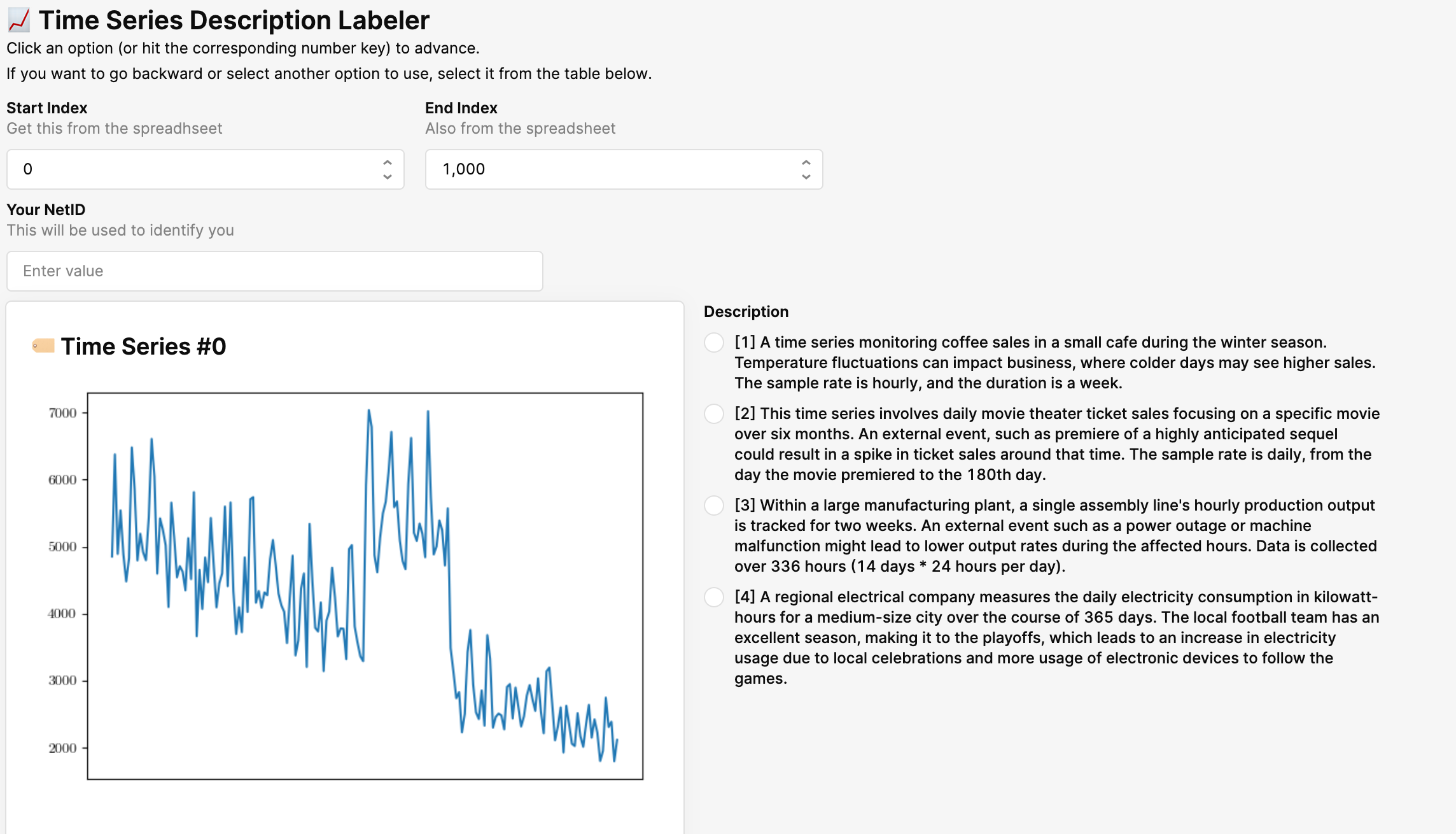}
    \caption{A screenshot of the tool used by human annotators in the etiological reasoning task (Section \ref{subsec:entailment})}
    \label{fig:annotatorTool}
\end{figure*}
}

\newcommand{\tableMainResults}
{
\begin{table}[t!]
\small
\centering

\begin{tabular}{lcccc}
\toprule
\textbf{Model/Task} & \textbf{Etiological Reasoning} & \multicolumn{3}{c}{\textbf{Question Answering}} \\
\cmidrule(lr){3-5}
&  & \textbf{One TS} & \textbf{Two TS} & \textbf{Perturbed} \\
\midrule
Random baseline	&	25\%   & 25\%   & 25\%    &  25\% \\
Human	&	66.1\%&	-	& 67.0\%	&  61.7\%\\
\midrule
\llamas	- No TS	&	N/A\textsuperscript{\textdagger} & 78.4\%	&	24.7\% & 25.6\%\\
\llamas	&	27.3\% &  78.8\%	&	25.2\%  & 24.3\%\\
\llamat	 - No TS	&	N/A\textsuperscript{\textdagger} & 82.6\%	&	26.3\% &    25.6\%\\
\llamat &	27.8\% &  82.5\%	&	25.8\%  &   25.6\%\\
\gptt - No TS	&	N/A\textsuperscript{\textdagger} &	90.4\%**	&	29.8\%**	&	26.3\%\\
\gptt  & 33.5\%** &   88.2\%**	&	27.4\%**	&  27.7\%\\
\gptf - No TS	&	N/A\textsuperscript{\textdagger} &	\textbf{92.6\%}* &	51.3\%*	&	28.4\%\\
\gptf & 33.5\%* &    92.3\%*	&	52.7\%*	&    28.4\%\\
\gptf-Vision &	33.5\%*&	91.8\%*	&	\textbf{53.6\%}*   & \textbf{30.5\%}\\
\midrule
\textbf{Gap - Human vs Best LM} &  32.6\% & - & 13.4\% & 33.3\%\\
\bottomrule
\multicolumn{5}{l}{\small  *\gptf generated all data and its performance should be interpreted with caution (Section~\ref{sec:dataset}).}\\
\multicolumn{5}{l}{\small **Since \gptt may share training data with \gptf, these concerns may transfer to \gptt.}\\
\multicolumn{5}{l}{\textdagger These results are not included for etiological reasoning because in this task models }\\
\multicolumn{5}{l}{\textit{only} have the time series (and no metadata) as input.}
\end{tabular}

\caption{Accuracy of LMs on Etiological Reasoning and Question Answering. The human performance was evaluated on randomly selected subset of data across both these tasks (N=500). \textbf{No TS} indicates that the model was evaluated without the time series as input (i.e. with only metadata in the prompt).  \textit{Etiological Reasoning:} LM performance is near-random for \llama models and slightly better for \gptt models. Human performances is significantly higher. \textit{Question Answering:} LM performance on single time series questions is high even when the time series is not given to the model and so we caution against interpreting these results as successful time series reasoning. When evaluating a related setting with two time series, LM performance drops to near-random for \llama models, and is slightly better for \gpt models, though again clearly trailing human performance (Section \ref{subsec:comparison}).
}
\label{tab:main_results}
\end{table}
}

%% file: 001_introduction.tex
\section{Introduction}

Time series measure how systems change over time and contain information that is uncommon in language. They are a critical data modality in healthcare~\citep{moridTimeSeriesPrediction2023}, finance~\citep{sezerFinancialTimeSeries2019}, agriculture~\citep{kamilarisDeepLearningAgriculture2018}, economics~\citep{nerlove2014analysis}, political science~\citep{beck2011modeling},
astronomy~\citep{bensonForecastingSolarCycle2020}, signal processing~\citep{jagannathRedefiningWirelessCommunication2021}, and beyond. As the scientific community races to bring language models (LMs) to these domains, we must ensure LMs can support decisions about these sources of valuable information. If successful, LMs could perform novel tasks like citing patterns and events in time series as evidence for observations and inferences, drawing interpretable conclusions from complex dynamical systems, or learning to recognize and respond to temporal patterns.
\figHero

Several recent works have shown that LMs can be used for zero-shot time series tasks, though nearly all focus on forecasting. These works typically forecast by structuring historical observations as raw text \citep{liuLargeLanguageModels2023, xuePromptCastNewPromptbased2023, zhangLargeLanguageModels2024, gruver2023large} or images \citep{li2023time}. This is promising work, and suggests language models may someday demonstrate the same remarkable zero-shot performance that they do with text and images. But it remains unknown whether non-trivial forecasting implies that LMs can \textit{reason} about time series, as opposed to simply generating matching temporal patterns that appear in their inputs. In fact, recent works indicate that a LM's ability to generate data does \textit{not} imply deeper reasoning \citep{west2023generative, hessel2022androids}.

In this work, we develop, apply, and release a framework to ultimately find that despite excitement about using LMs for time series analysis, \underline{\textbf{\smash{current language models are remarkably bad at zero-shot time series reasoning}}}. We propose three components of time series reasoning.  First, for a LM to reason about time series it must be able to consider the etiology (the set of possible causes) of a time series through \textbf{etiological reasoning} (Figure \ref{fig:hero}(a)). For example, given a time series of slowly rising freezer temperatures, a good model would hypothesize that this rise could have been caused by a power failure or an open freezer door. Second, a successful model should excel at \textbf{question answering} and be able to address queries about time series and how they relate to one another (Figure \ref{fig:hero}(b)). For example, given the time series of COVID transmission rates in two cities, a model should be able to identify which series most likely represents a lower overall mortality. Finally, time series reasoning implies \textbf{context-aided forecasting}, wherein a language model can leverage its world model and natural language context to aid in forecasting (Figure \ref{fig:hero}(c)). For example, if a language model is told that a negative news story will come out about a company, it should integrate this information into its prediction, potentially forecasting that its stock price will trend downward. 



To evaluate LMs we create a first-of-its kind dataset that contains 230k time series multiple choice questions and 8.7k pairs of synthetic time series and text captions that describe the series and the context in which it was observed (Section \ref{sec:dataset}). These data span a diverse set of time series scenarios across including health data, transport and traffic trends, finance, and more. 

We use this dataset to evaluate \textit{etiological reasoning} by tasking models to select the most probable time series caption given the observed time series (Section \ref{subsec:entailment}) and find that human annotators outperform language models by a margin of up to thirty percentage points, with otherwise strong language models like \gptf barely doing better than random chance. 
Then, we test models on a \textit{question answering} task by augmenting our dataset to include 230k question-answer pairs (Section \ref{subsec:comparison}). Again, we find that human annotators significantly outperform language models, indicating that language models have limited capacity to interpret the information in time series.  
Finally, we evaluate language models on a \textit{context-aided forecasting} task (Section \ref{subsec:constgen}). We find that even with text descriptions of what will happen in future, \gptf struggles to incorporate this information, resulting in negligible improvements over models without additional context. 
Taken as a whole these results indicate that despite modest time series forecasting ability, current language models fail to reason about these ubiquitous, critical data despite considerable human performance on the same tasks.

%% file: 003_definitions.tex
\section{Forms of Time Series Reasoning}
\label{sec:forms_of_reasoning}
Here we propose a rigorous (though non-exhaustive) definition of time series reasoning.

Consider a univariate uniformly-sampled time series of $n$ observations, $x = \{x_0 \cdots x_n\}$, $x \in \sR^n$. Suppose that an autoregressive language model $M$ is able to represent this time series as input and produce time series observations and text as outputs.\footnote{For models evaluated in our experiments (excluding \gptf-Vision, and the LLaVA and Whisper variants in Section \ref{subsec:training_models}) a language model represents a time series by casting its values into strings. Our definitions are intentionally agnostic to the model's input representation.} That is, $M$ estimates the probability $p$ of an output token sequence $Y$ given some context tokens $C$ and the time series: $p_M(Y |x,C) = M(Y,x,C)$.

\begin{definition}[Etiological Reasoning]
\label{def:etiological}
Etiological reasoning is the property by which language models are able to hypothesize about the cause of a time series. That is, given a time series $x$, textual instructions as context $C$, a correct description $D^+$ of how $x$ was generated and an incorrect description  $D^-$, a language model should assign higher probability to $D^+$:
\begin{align}
    p_M(D^+ |x,C) > p_M(D^- |x,C)
\end{align}
\end{definition}
Language models that can reason about time series should also be able to answer questions about the behavior and implications of a time series. 

\begin{definition}[Question Answering] We define question answering as a model's ability to use information in the time series $x$ to interpret queries about the time series or the events surrounding the scenario it represents. 

For the sake of evaluation, the questions should be time-series dependent---correct answers should be unattainable without interpreting $x$.
For example, given an ECG, a dependent question might be, \qt{Does this signal demonstrate atrial fibrillation?} while a trivially non-dependent question would be, \qt{Who was the first president of the United States?} Formally, given a question $Q$ and an answer $A^+$, the model should predict
\begin{align}
    p(A^+ |x, Q) \gg p(A^+|Q)
\end{align}
A language model should be able to exploit this information. In a multiple-choice setting, given a correct answer $A^+$ and an incorrect answer $A^-$:
\begin{align}
    p_M(A^+ |x, Q) > p_M(A^- |x, Q)
\end{align}
\end{definition} 

Finally, for an LM to reason about time series it should be able to integrate relevant information from text into forecasts about how the time series will behave in the future.

\begin{definition}[Context-Aided Forecasting]
 Context-aided forecasting is the property by which a language model can use additional outside information about a time series to guide its forecasts.  Given the first $t$ observations of a time series and a relevant text description $D$, the model should predict: 
\begin{align}
    p_M(x_{t+1} \cdots x_n | x_0 \cdots x_t, D ) >  p_M(x_{t+1} \cdots x_n | x_0 \cdots x_t)
\end{align}
Note that $D$ must provide some meaningful information about the behavior of $x$.
\end{definition}

%% file: 004_dataset.tex
\section{Dataset}
\label{sec:dataset}
Evaluating these forms of time series reasoning requires pairs of time series and highly-relevant text descriptions. Without a strong relationship between the two, it is impossible to determine if a model's failure to reason about time series is due to poor fundamental capabilities or a poorly-designed evaluation. However, there is no general corpus of time series and natural language descriptions that captures such relationships (Section \ref{subsec:related_work_datasets}). To address this challenge, here we contribute a first-of-its-kind dataset of synthetic multi-domain time series and highly relevant text captions.
\figScenario
We prompt GPT-4 to generate descriptions of environments that change over time alongside executable Python functions that generate corresponding time series. A naive solution is to generate a time series as text, however autoregressive language models struggle to generate text with long range interactions \citep{bubeck2023sparks} and demonstrate poor numerical reasoning \citep{Akhtar2023ExploringTN,dziriFaithFateLimits2023}.  Accordingly, time series that are generated as text exhibit poor coherence and are of overall low quality (Figure \ref{fig:TextCSCode}). Instead, we leverage recent language models' capacity to generate code\citep{Zhong2023CanCR,Chen2021EvaluatingLL, Wang2022ReCodeRE}. We therefore prompt \gptf to produce \textit{data generating functions} in the form of Python scripts. We ask the model to \qt{imagine a scenario} that would produce a time series. We then yield the following data for each scenario:
\begin{compactitem}
    \item A \textbf{caption} of the scenario that generated the time series.
    \item Five \textbf{characteristics} of a time series which matches this description.  
    \item A \textbf{generative function} which, when executed, returns the time series as an array.
    \item \textbf{Metadata} about the time series, including its start and end timestamp, its sample rate, units, a short caption of less than five words which summarizes the scenario. 
\end{compactitem}

To encourage diversity during generation, we append the latest twenty short descriptions to each new prompt and ask the model to a generate a scenario that is as distinct as possible from these previous generations. Empirically, this step is important for maintaining variety in the generated results. The full prompt is available in Section \ref{sec:prompt_for_scenario_generation}. Finally, we filter the scenarios by removing multivariate time series and those with complex, missing, or infinite values, resulting in 8.7k scenarios. Next, we feed 100 captions into \gptf and ask the model to categorize these time series into ten domains (Figure \ref{fig:Donut}). We then automatically apply these categories to the remaining 8.7k scenarios (Figure \ref{fig:OneOfEach}). We manually reviewed 50 scenarios and found no substantial inaccuracies between the captions and the time series.

To quantify the quality of these data, we evaluate human subjects' time series reasoning abilities. As discussed in Sections~\ref{subsec:entailment} and \ref{subsec:comparison}, human subjects achieve far above random performance and substantially outperform existing language models. This implies that there is enough information in the time series and prompts to facilitate significantly higher performance than LMs currently exhibit. 

We include ten randomly selected scenarios (one from each category) in Figure \ref{fig:OneOfEach}.

%% file: 005_experiments.tex
\tableMainResults
\section{Etiological Reasoning: Near Random Performance} 
\label{subsec:entailment}
By defining time series reasoning (Section \ref{sec:forms_of_reasoning}) and creating our first-of-its kind dataset of time series and associated captions (Section \ref{sec:dataset}) we can evaluate the capacity of LMs to reason about these ubiquitous data. Reasoning implies an ability to provide explanations for observed phenomena. In our context if a model can reason about a time series then it should be able to hypothesize about how that series was generated. For example, given a time series with a strong daily seasonality \qt{sunlight intensity} is a more likely description than \qt{Nvidia stock price since 1999.} 

We evaluate entailment by tasking an LLM to select the correct time series caption from a set of four, with three incorrect captions (Figure \ref{fig:hero}(a)). We sampled incorrect descriptions by randomly selecting three captions from the remainder of the dataset. To encourage the models to focus on the time series itself and not on metadata like the series' units or start and end timestamps we only provided the values of the time series. Time series were encoded into text using the method from \cite{gruver2023large}. Details on this method are available in Section \ref{subsec:numerical_tokenization}. For \gptf-Vision we plotted the time series using the same method as \cite{li2023time}. 

To confirm the quality of our ground truth labels and contextualize model performance we performed a human evaluation. Ten annotators with significant expertise in data science and time series modeling labeled an average of 50 examples each for a total of 500. Since the models were not provided with time series metadata annotators were shown a linegraph with the x and y axes labels removed for consistency (Figure \ref{fig:annotatorTool}). We note that skilled humans often struggle to interpret even simple time series plots \citep{albersTaskdrivenEvaluationAggregation2014}, and so human performance on this task may not represent the upper bound of possible performance.

Our results show that all models perform remarkably poorly relative to the human baseline (66.1\% accuracy), with some models performing at or near random chance (e.g \llama with 27.3\% accuracy) (Table \ref{tab:main_results}). GPT-4-Vision performs best (34.7\%) while still falling short of human performance by over 30 percentage points.  

A natural question is whether text is the correct way to represent a time series. To answer this, we also experimented with training existing multimodal models on our data and found similarly poor performance (Section \ref{subsec:training_models}).

\textbf{Taken as a whole, these results indicate that current zero-shot language models are poor judges of time series etiology.}

\section{Question Answering: Trailing Behind Human-Level Proficiency} \label{subsec:comparison}
A LM that can reason about time series should be able to answer questions about a time series and the implications of the scenario it describes. To properly evaluate this property it should not be possible to answer the questions \textit{without} the time series. This avoids misleading performance estimates observed in Visual Question Answering with models performing well even without the associated image~\citep{wang2023filling}. A good candidate for these questions are counterfactual \qt{what-if}-style queries that ask the LM to interpret how the time series might be different if its related scenario were changed. For example, given a time series of coffee shop sales over the course of a day with a peak at 2pm, a good \qt{what-if} question might be, \qt{If half as many customers visited the shop at noon, would the peak sales change?}

We evaluate this ability by solving Multiple Choice Questions (MCQs) with four options -- one correct and three incorrect. 

We first introduce an intuitive process for synthetically generating time series MCQs, and demonstrate that these questions do not appropriately evaluate time series reasoning, since LM performance is high even without the time series as input, in violation of our aforementioned requirement. We then improve over this first procedure by synthesizing questions about the difference between two time series, which empirically makes it harder for LMs to guess the right answer without attending to the time series as well.

Similar to Section \ref{subsec:entailment}, unless otherwise noted all text-based methods used the time series formatting approach from \cite{gruver2023large} (details in Section~\ref{subsec:numerical_tokenization}). In Section~\ref{sec:ts_as_prompts} we experiment with other input representations and show no meaningful difference in performance. Human performance was again assessed using a team of ten data scientists who annotated 500 time series plots using the same data (metadata, time series [as a plot], and the short description) as the LMs. 

\subsection{Questions About One Time Series}
\xhdr{`What-if' MCQs created for single time series were trivial to answer}
An intuitive approach to generate MCQs for time series is to prompt a LM to use the time series and associated scenarios and metadata from Section~\ref{sec:dataset} to generate questions and answers. We again use \gptf, as questions generated by other LMs were always answerable without the timeseries (Section \ref{sec:mcq_other_llms}). First, we prompt \gptf with the with all the information generated in Section~\ref{sec:dataset}, \ie time series, short caption, characteristics, generative function, and metadata, to generate a potential counterfactual 'what-if' scenario. Second, we prompt \gptf to generate questions around the original time-series and the possible changes due to 'what-if' scenarios and obtain 100k single time series MCQs (full prompt in Section~\ref{sec:prompt_for_singleTS}, and examples in Section~\ref{sec:sts_mcq}). 

In early experiments, we found that giving the LM access to the full caption consistently led to questions that were entirely dependent on the caption and did not reference the time series. Even after removing the caption from the question generating procedure, all LMs achieved 78-92\% accuracy \textit{without} using the time series, demonstrating that these questions did not necessitate time series reasoning (Table~\ref{tab:main_results}).

We further experimented with changing the order of options within MCQs, used prompts with different sets of time-series features, generative functions, metadata, and presented time series as plain text and as tokens using the procedure in \llmtime~\citep{gruver2023large}. However, none of these attempts produced MCQs that required the time series. 

We make the following observations:
(1) Performance overall was high, ranging from 78-92\% \textit{without the time series}. This creates a false impression of LM time series reasoning ability, when really the performance stems from parametric LM knowledge. 
(2) Since these data and questions were generated by \gptf, with \gptt potentially sharing training data and other components, it is less surprising that they are significantly better than \llama models. We therefore caution to interpret these results as a sign of generalizable time series reasoning ability, which is further called into question by the experiments described next.

Since LMs performed well even in the absence of time series, we deemed this setting unsuitable for evaluating time series reasoning, and did not perform additional human evaluation.

\subsection{Questions About Two Time Series}
\xhdr{MCQs created using two time series led to near-random performance for all LMs (except the one generating the MCQs)}
To create time series MCQs that cannot be answered by LMs without attending to the time series itself, we consider another setting in which we first create 'what-if' scenarios for a time-series \textit{alongside a second time series that materializes this counterfactual scenario.} 
We create these MCQs using a three-step procedure.
\begin{compactitem}
    \item For each time series $x$ (Section~\ref{sec:dataset}) and a 'what-if' scenario as described in the previous paragraph, we query \gptf to produce the corresponding generative function that simulates a second time series, $\overline{x}$, that reflects the 'what-if' scenario.
    \item We use the 'what-if' scenario, short captions,  both time series $x$ and $\overline{x}$, and their generating functions to generate MCQs about similarities and differences between $x$ and $\overline{x}$.
    \item To ensure that all MCQs are answerable only in the presence of \textit{both} time series, we filtered out questions that \gptt could answer in the absence of \textit{any} time series, which led to almost half of the MCQs being discarded. In total, this process generated over 130k MCQs, with one correct and three incorrect answers each. An example of these questions is in Figure~\ref{fig:hero}.
\end{compactitem}

We make the following observations: 
(1) Relative to the single time series MCQs described in the previous section, all LMs, other than \gptf, decreased to close to random performance (Table~\ref{tab:main_results}).
(2) Only \gptf achieves non-trivial performance on this MCQ task. However, performance does not meaningfully increase when the time series is added to the LM input. Again, the fact that \gptf, with and without time series, achieves non-trivial performance may be because \gptf was used to generate these scenarios. We describe below an additional experiment that is consistent with this interpretation. 
(3) Human performance, when given the exact same information as the LMs is significantly higher than all LMs at 67\%  which perform at near-random performance (other than the aforementioned \gptf and \gptt exceptions). This gap demonstrates that higher performance should be possible for LMs.

One potential reason for LMs performing just as badly even with a time series representation is that these time series may not contain any relevant information. However, since human performance is substantial at 67\% we can rule out this possibility. The only model achieving meaningful levels of performance in the MCQ task with multiple time series is \gptf, and we have to caution again that \gptf was used to generate these MCQs and this evaluation is likely to overestimate generalization performance of \gptf. 


\subsection{Manually-Perturbed MCQs}
\xhdr{Minor manual perturbations in MCQs eradicate above-random zero-shot performance for
any LM, including GPT-4 which generated all data} Upon first inspection it is notable that \gptf achieved non-trivial levels of performance in question answering. However, we show that this performance is possibly explained by \gptf being the model used to synthetically generate these data and MCQ tasks, casting significant doubt on any actual time series reasoning ability of \gptf, and therefore \textit{all} of the LMs evaluated in this study. We demonstrate this by taking 144 samples from the previously described \qt{two time series} MCQ dataset and make manual perturbations to the answers. Concretely, for each question we select the correct answer for the MCQ and create a similar incorrect answer as a \textit{distractor} by editing the numerical values so that they are similar while still incorrect. We provide an example in Section~\ref{sec:adv_perturbed}.

In addition, we create a small set of 52 manually generated MCQs on non-synthetic real-world time series as a second dataset to evaluate the generalizability of any non-trivial performance observed thus far.
Specifically, we selected time-series examples from yearly unemployment rates in the USA, annual imports in the USA from China, and COVID-19 cases in Massachusetts, among others, and wrote associated MCQs (Section~\ref{sec:handcrafted}).

We make the following observations:
(1) Prior to the manual perturbations, \gptf and \gptf- No TS answered over half the MCQs correctly. However, after only minor changes to MCQ options performance decreases to near-random performance as well (Table~\ref{tab:main_results}). This strongly suggests that \gptf's above-random performance in all prior time series MCQ tasks is due to the fact that it created the data and MCQs itself, and that does not generalize to slightly varied settings. We hypothesize (i.e., do not claim or prove) that the prior non-trivial performance is explained by the model recognizing likely correct answers due to artifacts of the distribution that this LM models. 
(2) In the manually created real-world dataset, \gptf's performance is significantly lower than observed in synthetic data (36.6\%) while still better than random chance. However these real-world time series were collected from online datasets and represent world knowledge that could be part of the LM's parametric knowledge, and not indicate genuine zero shot-time series reasoning. This is also supported by the fact that \gptf with and without access to the actual time series again perform similarly.



In summary we show that LMs exhibit (near-)random performance on meaningful QA tasks while human evaluations demonstrate that significantly better performance is possible, using the exact same set of information given to the LMs. \textbf{In none of these zero-shot evaluations did LMs perform better with than without the time series, suggesting that current LMs cannot perform time series reasoning.}

\section{Context-Aided Forecasting}
\label{subsec:constgen}
We next evaluate whether capable LMs can leverage relevant textual context when forecasting future time series values.
We build on recent works that find LMs can non-trivially zero-shot forecast time series ~\citep{gruver2023large,xuePromptCastNewPromptbased2023}.
Using the same zero-shot forecasting method as \llmtime \citep{gruver2023large}, we experiment with prepending different corresponding textual context alongside the time series.
Specifically, we randomly select 2000 time series with their captions, descriptions, and metadata, feed the first 80\% of the time series into GPT-4 and then forecasts the remaining 20\% of the timesteps. Further method details are in Appendix \ref{app:num_token}.  This textual context contains highly-relevant information, occasionally including \textit{future information} about the series' behavior.
To understand how well these methods compare to a simple baseline we include the \qt{Predict Median} baseline, which simply computes the median of the first 80\% of a time series' values then repeats it for the forecasting window. 

We measure forecasting success using the common metrics Mean Absolute Error (MAE) and Mean Squared Error (MSE). Since the values of the time series in our dataset span several orders of magnitude we min/max and z-score normalize values before computing these metrics so that error on high-magnitude series does not dominate perceived model performance.

\begin{figure}[t]
    \centering
    \hfill
    \subfigure[MAE with max-min normalization]{\includegraphics[height=3cm,width=0.48\textwidth]{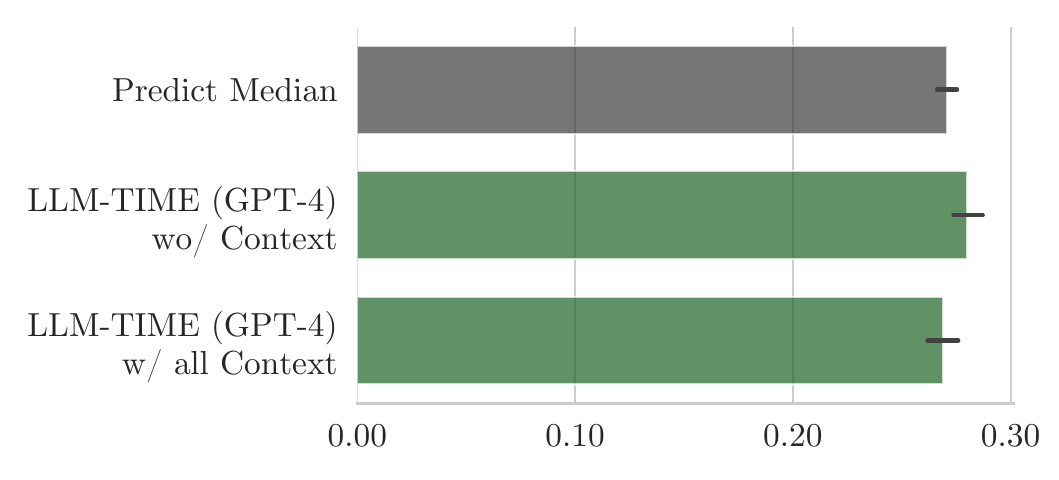}}
    \hfill
    \subfigure[MAE with z-score normalization]{\includegraphics[height=3cm,width=0.48\textwidth]{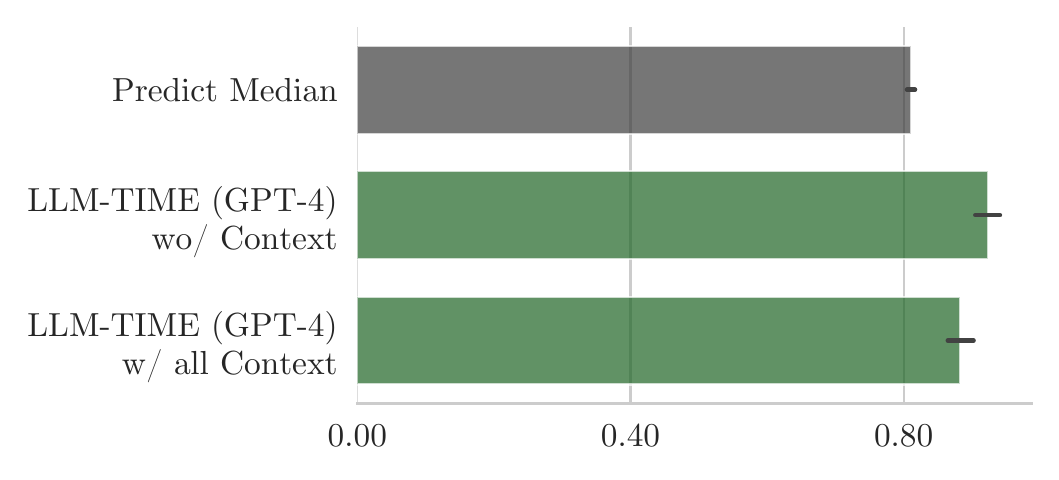}}
    \caption{After adding contextual information corresponding to the time series, forecasting performance improved \textit{marginally} and is still the same or worse than a simple baseline that only predicts the median of the historical signal (Section\ref{subsec:constgen}).}
     \label{fig:gpt4-forcast_simple}
\end{figure}

\begin{figure*}
   \centering
    \includegraphics[width=0.98\linewidth]{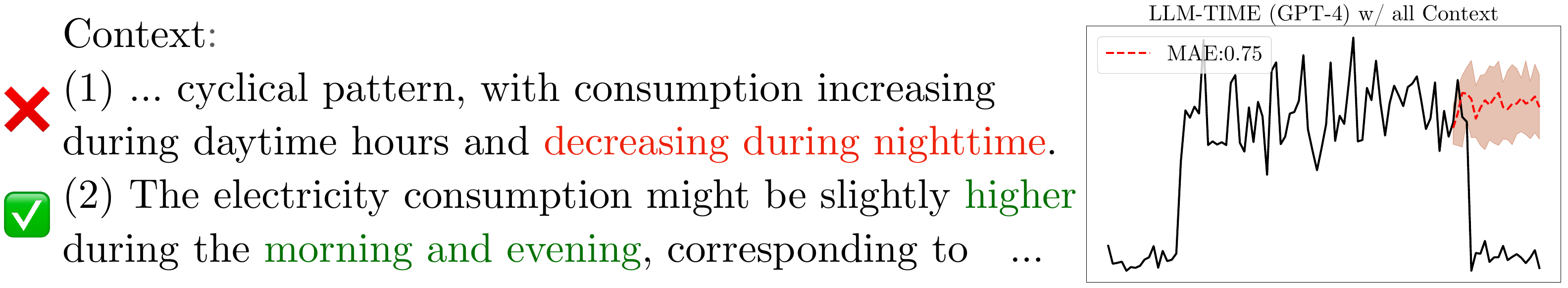}
    \caption{An example of forecasting with context. This data is sampled every 15 minutes from 0:00 to 23:45, with electricity usage dropping sharply near midnight. The starting time for forecasting is 19:15. The left side displays the captions in our dataset and the right side presents the performance of \llmtime (GPT-4) with all context which fails to incorporate this highly-relevant information.}
    \label{fig:hurts_reasoning}
\end{figure*}

\xhdr{Highly-relevant captions barely change LM forecasts}
As shown in Figure~\ref{fig:gpt4-forcast_simple}, adding all textual context barely changes MAE, despite often having access to descriptions of future information. 
Of 2,000 zero-shot samples, only 1,040 show improvement in MAE when the full context is shown and in the remaining time series MAE \textit{increases}. An example is illustrated in Figure~\ref{fig:hurts_reasoning}, showing that the LM ignores potentially useful information in the context. We also experimented with other combinations of metadata, characteristics, and descriptions and found that adding more information gradually improves performance, but overall performance remains below or comparable to the weak \qt{Predict Median} baseline (Section~\ref{subsec:forecast_res}).

This lack of improvement is surprising and demonstrates a clear gap in these LM-powered methods' capacities to leverage relevant text when forecasting time series.
Further, neither LM-power forecasting method clearly outperforms the simple \qt{Median Prediction} baseline. We note that because our series were intentionally designed to contain interruptions from external events (Section \ref{sec:dataset}) median prediction is a particularly weak baseline on our dataset. 

\textbf{This experiment shows that current LMs largely fail to use context to inform forecasting.}

%% file: 002_related_work.tex
\section{Related Work} 

\subsection{Datasets for Time Series and Language} \label{subsec:related_work_datasets}
There are dozens of prominent time series classification and forecasting datasets, many of which aggregate data from diverse domains \citep{tan2020monash,ucrarchive2018, Bauer2021LibraAB, Grauman2023EgoExo4DUS}. Unlike these datasets, which focus exclusively on time series, our goal is to evaluate the relationship between time series and text and motivate time series reasoning as an area of research beyond forecasting and classification. Some datasets focus on single-domain question answering with time series. \citet{Oh2023ECGQAAC} and \citet{Xing2021DeepSQAUS} provide a question answering dataset based on templated questions relating to ECG features and activity recogntion, whereas \citet{Xie2023PIXIUAL} present templated questions that concern tweets and historical stock price data. 

\subsection{Language Models and Time-Series}
Recent works have demonstrated that language models (LMs) perform well in time series tasks, such as forecasting~\citep{gruver2023large} or classification~\citep{zhou2023fits}. These can be categorized into two paradigms. The first involves fine-tuning language models, such as Bert or \llamas, for specific tasks and datasets~\citep{zhou2023fits,jin2024timellm,cao2023tempo}. The second approach entails inputting specially tokenized time series into an LLM for forecasting , imputation, and classification tasks~\citep{gruver2023large,xuePromptCastNewPromptbased2023}.



Most tasks that use textual context to aid time series forecasting focus on a single domain and require fine-tuning the model itself with domain-specific data. In cross-domain tasks, the strategy often involves fitting one dataset and then transferring to another~\citep{jin2024timellm,cao2023tempo,zhou2023fits,wang2023promptbased}. This approach is not suitable for our dataset, where each time series originates from a different setting, making it impossible to fit each domain individually or have sufficiently many similar sequences for transfer. Therefore, to evaluate our entirely cross-domain dataset, we utilize the latest state-of-the-art zero-shot method, \llmtime \citet{gruver2023large}, as our baseline.



%% file: 006_conclusion.tex
\section{Conclusion}
We identified three forms of time series reasoning and used them to create a first-of-its-kind dataset of time series and highly relevant text. We then used this dataset to assess etiological reasoning, question answering, and context-aided forecasting. Given the substantial gap between language model and human performance on the first two tasks, and mediocre performance on the third, we identified opportunities for the NLP community to develop models that can deeply reason about these critical data. 

%% file: 100-appendix.tex
\section{Appendix}
\figTextVSCode
\subsection{Numerical Tokenization}\label{app:num_token}
\label{subsec:numerical_tokenization}
We use the \llmtime~\citep{gruver2023large} as a baseline for "contextual reasoning" to evaluate LLM`s reasoning performance in time series forecasting when captions are provided. The performance of \llmtime is partly attributable to their special numerical tokenization method. The original input ($20.88,\: 20.20,\:20.48, \ldots$ below), is first z-score normalized and then scaled to a constant power of ten ($1e3$ below):
\[
20.88,\; 20.20,\; 20.48, \ldots \rightarrow 1.0522,\; 1.0178,\; 1.0324, \ldots \rightarrow 1052,\; 1017,\; 1032, \ldots
\]

Note that there are subtle differences in tokenization for GPT-3 and LLama.\footnote{https://github.com/ngruver/llmtime}

\section{Additional Results}
\label{app: additional}

\subsection{Training Multimodal Models on Etiological Reasoning Task}
\label{subsec:training_models}
Is putting time series into a prompt as text the best way to model these data? Here we experiment with five alternative modeling techniques, each adapted from an existing multimodal architecture. When training models owe wanted to keep the results roughly comparable to zero-shot experiments so we reserved the \qt{Health and Medical Data}, \qt{Agricultural and Food Production} and \qt{Educational and Public Services} categories for testing and trained on the remainder. 

\xhdr{Whisper} Speech-to-text models can be thought of as special cases of time-series-to-text models since microphone-recorded audio is a 1D sensor reading. We modify Whisper \citep{radford2022robust} to compute spectrograms of arbitrary time series and fuse these with GPT-2 inputs via cross attention. 

\xhdr{\llava-Matplotlib-Zero-Shot} \citep{liu2023visual} supports visual instruction tuning by training a linear adapter between a vision encoder and a language model's token embedding space. Following \citet{li2023time} we encode time series by plotting them in Matplotlib and saving the results as 224x224 images. These images are fed directly into LLaVA's pretrained CLIP encoder. As the name suggests, this model was not trained and instead relies entirely on the pretrained LLaVA weights.

\xhdr{\llava-Matplotlib} This experiment is the same as the previous, but we began by tuning LLaVA's adapters using the seven held-out scenario categories. 

\xhdr{\llava-Spectrogram} Spectrograms are 2D representations of a time series and can be passed to standard vision encoder. For this experiment we computed spectrograms and fed them imto LLaVA's clip encoder. 

\xhdr{\llava-TimesNet} In this experiment we replaced LLaVA's CLIP encoder with the TimesNet \cite{wuTimesNetTemporal2DVariation2023} encoder. TimesNet adaptively maps 1D time series signals into a 2D space that can be interpreted by computer vision kernels and was designed as a general-purpose time series encoder. Since there is no pretrained TimesNet checkpoint in this experiment we freeze only the LLaMA backbone and allow the model to learn weights in the encoder. 

The results show that all models struggle to learn etiological relationships between time series and text. Each model performs within an epsilon of random performance (25\%). We conclude that even models finetuned on these data have limited capacity to reason about time series.

\begin{table}
\small
\centering

\begin{tabular}{lccc}
\toprule
\textbf{Model/Task} & \textbf{Etiological Reasoning} \\
\midrule
Human	& 66.1\% \\
\midrule
Whisper	&	23.6\%	\\
\llava-Matplotlib-Zero-Shot	&	24.3\%	\\
\llava-Matplotlib	&	26.1\%\\
\llava-TimesNet	&	23.5\%\\
\llava-Spectrogram	&	26.1\%\\
\bottomrule
\end{tabular}
\caption{Perfomance of multimodal models trained on the etiological reasoning task (Section~\ref{subsec:entailment})}
\label{tab:additional_etiological_results}
\end{table}

\figOneOfEach

\subsection{MCQ Generation using other LMs}\label{sec:mcq_other_llms}
Here, we evaluate the ability of LM other than \gptf to generate MCQs. Specifically, we created counterfactual scenarios and the corresponding questions using two LM -- \llamat and \gemma~\cite{gemma_2024}. Across each setting, we used 100 time series examples and created a set of almost 1000 MCQs for each LLM. The results across these datasets clearly show that \gptf achieves significant performance across the MCQs generated using \llamat and \gemma, even in the absence of any time series information (Table~\ref{tab:other_llm_results}). This can be attributed to the limited ability of LMs in understanding the dynamics within time series data and creating questions solely based on their textual descriptions. These results reinforce that other LMs may not be suitable for generating time series-specific questions and, consequently, for training models to evaluate time series reasoning ability.

\begin{table}[]
\label{}
\small
\centering
\begin{tabular}{lcc}
\toprule
\textbf{Model/Generator LM} & \textbf{\llamat} & \textbf{\gemma}\\
\midrule
\llamat	- No TS &	88.1\% & 88.5\% \\
\llamat & 87.0\%	& 87.3\% \\
\gemma - No TS	& 86.6\%	& 88.5\% \\
\gemma	& 87.2\% & 88.3\% \\
\gptt - No TS	& 96.8\% & 97.0\% \\
\gptt  & 96.4\% & 97.1\% \\
\gptf - No TS  & 97.5\% & 97.7\% \\
\gptf & 97.2\% & 97.4\% \\
\bottomrule
\end{tabular}
\caption{Accuracy of LMs on counterfactual MCQs generated using \llamat and \gemma.}
\label{tab:other_llm_results}
\end{table}

\subsection{Using Different Methods to Prompt Time-series} \label{sec:ts_as_prompts}
Here, we evaluate different methods of passing a time series to a language model. This task is incredibly important, as recent research has shown that changing the tokenization for time series can lead to it being easily confused by language models and can result in state-of-the-art results in forecasting~\cite{gruver2023large}. Therefore, in this section, we compare two methods used in \llmtime~\cite{gruver2023large}: specifically, passing tokens as comma-separated values and using the tokenization procedure described in Appendix~\ref{app:num_token}. Our results across both methods show insignificant differences in the ability of LMs to answer MCQs (Table~\ref{tab:llm_diff_ts}). However, we note that LM with time series encoded as \llmtime obtains slightly better performance.

\begin{table}[]
\small
\centering
\begin{tabular}{lcccc}
\toprule
\textbf{Model/Task} & \multicolumn{2}{c}{\textbf{Single TS MCQ}}  & \multicolumn{2}{c}{\textbf{Multiple TS MCQ}} \\
\cmidrule{2-5}
 & \textbf{Plain Text} & \textbf{\llmtime} & \textbf{Plain Text} & \textbf{\llmtime}\\
\midrule
\llamas	& 78.6 & \textbf{78.8\%} & \textbf{25.2\%} & 25.1\%\\
\llamat & 82.4 & \textbf{82.5\%} & 25.7\% & \textbf{25.8\%}\\
\gptt  & 88.2 & 88.2\% & 27.0\% & \textbf{27.1\%}\\
\gptf & 92.2 & \textbf{92.3\%} & 52.5\% & 52.5\%\\
\bottomrule
\end{tabular}
\caption{LMs' accuracy on MCQs when time-series are given as comma-separated values in plain text and tokenized using \llmtime.}
\label{tab:llm_diff_ts}
\end{table}

\subsection{Examples of Single Time Series MCQs} \label{sec:sts_mcq}
Here we provide a few examples of single-time series MCQs. Specifically, for the time series given in Figure~\ref{fig:singlets}, we queried \gptf and obtained the following MCQs.
\begin{lstlisting}
Q. How would the series be affected if the cafe started to remain open all night?
    A. If the cafe started to remain open all night the timeseries <@{\nolinebreak}@>would show no change in customer counts as the patterns remain the same.
    <@\textcolor{teal}{B. If the cafe remained open all night the periods that previously showed zero customer counts due to closure would now show some level of customer activity. However the counts during these late hours would typically be lower compared to the breakfast and lunch times.}@>
    C. If the cafe started to <@{\nolinebreak}@>remain open all night the timeseries would show higher customer counts during the day and a decrease in counts during the night when the cafe is closed.
    D. If the cafe started to remain open all night the timeseries <@{\nolinebreak}@>would show significant spikes in customer counts throughout the day.

Q. How would the time series <@{\nolinebreak}@>be different if the TV show started to air on Wednesdays instead of Sundays?
    A. If the TV show started to air on Wednesdays instead of Sundays the timeseries would show no change in customer counts as the patterns remain the same.
    <@\textcolor{teal}{B. If the TV show aired on Wednesdays instead of Sundays the pronounced spikes in the customer counts would shift to reflect this change. This means we would start to see the spikes on Wednesdays and continue for the next few days following the broadcast.}@>
    C. If the TV show started to air on Wednesdays instead of Sundays the timeseries would show increased customer counts throughout the week.
    D. If the <@{\nolinebreak}@>TV show started to air on Wednesdays instead of Sundays the timeseries would show a decrease in customer counts on Wednesdays and an increase on Sundays.,
Q. What would the effect on the customer count <@{\nolinebreak}@>be if the cafe started serving dinner and remained busier during evening hours?
    A. If the cafe started serving dinner and remained busier during evening hours the timeseries would show a decrease in customer counts during dinner time.
    B. If the cafe started serving dinner and remained busier during evening hours the timeseries would show increased customer counts only during dinner time.
    C. If the cafe <@{\nolinebreak}@>started serving dinner and remained busier during evening hours the timeseries would show no change in customer counts as the patterns remain the same.
    <@\textcolor{teal}{D. If the cafe became busier during dinner time then the unknown counts during evening hours would increase. This could introduce another cyclical pattern in the time series corresponding to dinner hours similar to those observed during breakfast and lunch times.}@>
\end{lstlisting}
\figsinglets

\subsection{Manually Perturbed MCQs} \label{sec:adv_perturbed}
In this section, we highlight the procedure we use to manually perturb the MCQs generated by \gptf. In detail, we aim to test the robustness of \gptf across slightly modified versions of the same set of MCQs it generated. For this, consider the following MCQs generated by \gptf for two independent time series. These questions aim to compare the time series updated by the 'what-if' scenario with the original time series.
\begin{lstlisting}
Q. Do both the <@{\nolinebreak}@>original and updated time series have the same starting base level of pollution?
    A. No the updated series starts at 0 units of pollution.
    B. No the base level in the updated series is 1500 units.
    <@\textcolor{teal}{C. Yes both start with a base level of 1000 units of pollution.}@>
    D. No the base level in the original series is 500 units.

Q. Is there a change in the visitor <@{\nolinebreak}@>count during the pandemic period in the modified time series compared to the original?
    A. The visitor count during the pandemic does not drop to 0 in the modified series.
    <@\textcolor{teal}{B. There is no change the visitor count during the pandemic period drops to 0 in both.}@>
    C. The visitor count during the pandemic becomes 1500 in the modification.
    D. The pandemic period is removed in the modification.
\end{lstlisting}

To change the question, we select the correct option -- option C and option B respectively, and create a similarly looking incorrect option. Later, we replace this perturbed option with a randomly selected incorrect option and test the LMs' ability in responding to the MCQ. The following shows the updated MCQs with options D and C being the perturbed options. Upon evaluating both the MCQs, we note that \gptf and other LMs selected the perturbed option as their choice of answer. However, we also note that the LMs across different runs selected the correct option, \ie option C and Option B too. But the goal of the manual perturbation succeeds in showing that LMs cannot understand and select an answer using a time series and mostly select options based on their similarity to the option they originally generated.

\begin{lstlisting}

Q. Do both the <@{\nolinebreak}@>original and updated time series have the same starting base level of pollution?
    A. No the updated series starts at 0 units of pollution.
    B. No the base level in the updated series is 1500 units.
    <@\textcolor{teal}{C. Yes both start with a base level of 1000 units of pollution.}@>
    <@\textcolor{red}{D. Yes both start with a base level of 500 units of pollution.}@>

Q. Is there a change in the visitor count <@{\nolinebreak}@>during the pandemic period in the modified time series compared to the original?
    A. The visitor count during the pandemic does not drop to 0 in the modified series.
    <@\textcolor{teal}{B. There is no change the visitor count during the pandemic period drops to 0 in both.}@>
    <@\textcolor{red}{C. There is no change the visitor count during the pandemic period drops to 10 in both.}@>
    D. The pandemic period is removed in the modification.
\end{lstlisting}

\subsection{Handcrafted MCQs} \label{sec:handcrafted}
Here we provide some examples of the completely handcrafted MCQs generated for real-world time series. Specifically, for the time series depicted in Figure~\ref{fig:covid}, illustrating the COVID-19 cases in the state of Massachusetts\footnote{https://www.mass.gov/info-details/covid-19-reporting}, we created the following questions.
\figreal
\begin{lstlisting}
Q. When was the number of cases the lowest in the time series?
    <@\textcolor{teal}{A. July 2, 2023.}@>
    B. June 30, 2024.
    C. August 1, 2024.
    D. November 30, 2023.

Q. When was the number of cases the highest in the time series?
    A. September 31, 2023.
    <@\textcolor{teal}{B. January 1, 2024.}@>
    C. March 2, 2024.
    D. August 30, 2023.

Q. What is the visual representation of the time series?
    A. The cases steadily decrease over time with no discernible peaks.
    B. The cases <@{\nolinebreak}@>show minor fluctuations that go down but then go up with an exponential rate.
    <@\textcolor{teal}{C. A first peak is obtained then a drop in cases. Then a second peak is achieved that is higher than the first.}@>
    D. The number of cases remains constant throughout <@{\nolinebreak}@>the observed period, however, they suddeny increase around the mid of March.

Q. What is the <@{\nolinebreak}@>difference between the max of both the peaks in September 2023 and January 2024?
    A. The <@{\nolinebreak}@>difference is negligible between the peaks of September 2023 and January 2024.
    B. September 2023 had a higher peak compared to January 2024.
    C. The peaks in September 2023 and January 2024 are identical.
    <@\textcolor{teal}{D. January 2024 is almost twice of September.}@>
\end{lstlisting}

\section{Prompt For Scenario Generation} \label{sec:prompt_for_scenario_generation}
We used the following prompt to generate the time series scenarios described in Section \ref{sec:dataset}.
\begin{lstlisting}
1.   Describe a scenario that might <@{\nolinebreak}@>produce a time series.  This scenario should include an external event and how it might influence the reading. Be sure to describe the sample rate of the time series and the duration over which it is sampled. The <@{\nolinebreak}@>description should be less than 100 words in length. Delimit this description with the XML tag <description>. 

    The time series must be less than 1000 observations in length, be a <@{\nolinebreak}@>single variable, have no values greater than 1e6, and have no missing values. 

    Also add a summary of the description, no more than 25 words <@{\nolinebreak}@>in length with the tag <description_short>. Also add  summary, no more than three words in length with the tag <description_tiny>. The scenario should be as different as possible from any of the following: [<previous_descriptions>]

2.   You will generate a list of up to five characteristics of this specific time series, including patterns that you might expect to see in the series and how external events might cause distribution shifts in the data generating process. Delimit these characteristics with the XML tag  <characteristics>.

3.   You will write a <@{\nolinebreak}@>numpy  function called `generate_series` that takes no arguments and outputs a time series that matches the description.  All parameters from the data generating process should be drawn from reasonable distributions. The function must return a single numpy array.  Place this code inside a python markdown block and delimit your code with the XML tag <generator>. Do not call the function, simply define it. You should also make sure that the scale of time series is realistic. For example, a time series of a quantity like stock price should never be less than zero. 
    
4.   Return a json string, delimited by the tag <metadata<@{\nolinebreak}@>> that contains the units of the time series and the timestamps corresponding to the first and last values. Remember that in JSON format datetimes must be passed as strings. Also include a string that relects the frequency of the time series.

Here is an example of a complete response: 
<description> *your description* </description> 
<description_short> *your description* </description_short>
<description_tiny> *your description* </description_tiny>
<characteristics> *your characteristics* </characteristics>
<generator> 
    ```python
    def generate_series():
        # your code here
        return x
    ```
</generator>
<metadata>        
        {
        "start": x,
        "end": y,
        "units": z,
        "frequency" : freq
        } 
</metadata>
\end{lstlisting}

\section{Prompt For MCQ Generation}
\subsection{Prompt for Single Time-Series MCQs} \label{sec:prompt_for_singleTS}
We use the following prompt to generate the MCQs around single-time series described in Section~\ref{subsec:comparison}.
\begin{lstlisting}
1.   Given <@{\nolinebreak}@>a description of a time-series, a set of sentences describing its characteristics, and a python code segment that generates this time-series. You have to create five counterfactual question-answer pairs. Counterfactual reasoning questions involve exploring hypothetical scenarios by considering what would have happened if certain events or conditions had been different from what actually occurred. 
2.   For example, 'What will the time-series look like if some event occured?'. Generate a wide-range of questions. Create questions and answers that avoid referencing or directly quoting code or the description. Avoid asking questions specifically tied to the description or the Python code. The questions should require an understanding of time-series dynamics for accurate answers. 

3.   The answers should not mention the <@{\nolinebreak}@>description or the code at all. Provide the questions and answers in the following exact format: '{'category':'"+et+"', 'question':'', 'answer':''}'. Ensure that each question and its corresponding answer are presented on the same line, with each new question starting on a new line for a clear and organized format. 

4.   Using the set of question-answer <@{\nolinebreak}@>pairs, create three incorrect answer options for each question. Your incorrect answers should have similar lengths compared to the correct answers. The input format is: '{'question':'', 'answer':''}'. In the output, you should copy the question and answers from the input and provide incorrect options in the following format: '{'question':'', 'answer':'', 'incorrect answer 1':'', 'incorrect answer 2':'', 'incorrect answer 3':''}\n'. Each new question should start on a new line. Do not separate question, its answer and options into different lines. Ensure that each question, its corresponding answer and incorrect answers are presented on the same line.   Do not use any double quotations within the text. 

5.   Avoid the use of contractions in all kinds of notations. Instead, use the full forms for greater clarity. If there exists any contraction in the question or answer, then replace it with the full-form. Do not generate any additional text.
\end{lstlisting}

\subsection{Prompt for Multiple Time-Series MCQs} \label{sec:prompt_for_multipleTS}
For generating MCQs that operate at the intersection of multiple time-series, we employed the following steps:
\subsubsection{Creating a list of 'what-if' scenarios for a time series}
\begin{lstlisting}
1.   You have been given a description <@{\nolinebreak}@>of a time series and a code that generates the time-series. Your task is to create five counterfactual questions that someone can ask regarding this time series. 

2.   Try to formulate questions that are distinct from <@{\nolinebreak}@>each other. Additionally, ensure that the questions aim to bring about significant changes to the time series. Make sure that the new time series can be easily generated by modifying the code and do not ask extremely difficult questions. 

3.   Format the <@{\nolinebreak}@>output as follows:'{'question'}'\n, with each new question starting on a new line. The counterfactual questions should explore hypothetical scenarios and involve 'What-if' type inquiries. The questions should not include values directly from the original time series or code. For instance, 'What if the start was 25 units' is preferred over 'What if the start was 25 units instead of 20 units?'. 

4.   Avoid referencing random noise, the random number generator<@{\nolinebreak}@>, its mean, or variance in any question. Do not generate any additional text. 
\end{lstlisting}

\subsubsection{Creating a new time-series}
For each time series $x$ (Section~\ref{sec:dataset}) and a 'what-if' scenario outlined in the previous paragraph, we employ \gptf to generate the corresponding generative function. This function simulates a second time series, denoted as $\overline{x}$, reflecting the 'what-if' scenario. We used the following prompt to generate the updated time series

\begin{lstlisting}
1.   Generate a new Python code for a time series based on the <@{\nolinebreak}@>given code and description. The user will specify a change in the time series, and you should produce the updated code using the function name 'generate_series'. 

2.   Always ensure the length of the time series remains unchanged<@{\nolinebreak}@>. This is hard constraint that should not be violated. Keep realistic expectations and ensure the length of the time series remains unchanged. For example, (1) keep the rate of change consistent rather than the actual values. (2) Understand what changes the user's suggestion can make to the time series and then update the code accordingly. (3) Given a time series code, you have the freedom, and in some cases, the obligation, to modify any pre-defined maximum or minimum values specified in the original code to accurately represent the desired change. 

3.   Ensure that the new time <@{\nolinebreak}@>series adheres to real-world principles; for instance, maintaining a consistent rate of change under typical conditions. If the change demands that the time series has an offset by some units, then modify this value in the code as well.

4.   Return the output <@{\nolinebreak}@>in the format ```new code```, where the 'new code' is replaced by the updated code. Try to create code that generates a time-series that is significantly different from the time-series produced by the original code, but with same lengths. 

5.   Always return the code in a format <@{\nolinebreak}@>that can be executed directly using the exec() function. Avoid additional text.
\end{lstlisting}

\subsection{Creating MCQs}
Utilizing the 'what-if' scenario, brief captions, and both time series $x$ and $\overline{x}$, along with their generating functions, we construct multiple-choice questions (MCQs). These MCQs aim to evaluate the similarities and differences between the two time series. We used the following prompt to generate the MCQs around single-time series described in Section~\ref{subsec:comparison}.

\begin{lstlisting}
1.   Given two Python codes for generating time series<@{\nolinebreak}@>, the first representing the original time series with a description, and the second presenting a modification of the original time series under specific conditions. 

2.   Your task is <@{\nolinebreak}@>to ask five questions regarding the differences between both time series. Also ask five questions regarding the simmilarity between both time series. Additionally, provide answers to all the questions and three negative or incorrect options. Ask questions regarding the patterns within both time-series, such as how they appear, the rates of change, and any specific differences in trends. Format the output as follows:'{'category':'difference/simmilarity', 'question':'', 'answer':'', 'incorrect answer 1':'', 'incorrect answer 2':'', 'incorrect answer 3':''}'. 

3.   Make sure you follow the following rules: (1) Do not ask <@{\nolinebreak}@>question regarding the lengths or the number of data-points within both the time-series. (2) Ensure that the questions and answers give the impression of being created independently, in the absence of the code, solely by examining the time series. (3) Do not mention anything regrarding the random noise or random number generator in both the answers and questions. (4) Try to keep the answers short and not very detailed. (5) Ensure that each question and its corresponding answer are presented on the same line, with each new question starting on a new line for a clear and organized format. (6) Try to add numerical values to answers wherever possible, but make sure you use words such as 'seems to be' or 'around value' so that they appear to be approximate. Avoid unnecessary text and focus on precision.
\end{lstlisting}

\section{Additional Results for Context-Aided Forecasting}\label{subsec:forecast_res}

\begin{figure}[htbp]
    \centering
    \hfill
    \subfigure[MAE with max-min normalization]{\includegraphics[height=4cm,width=0.48\textwidth]{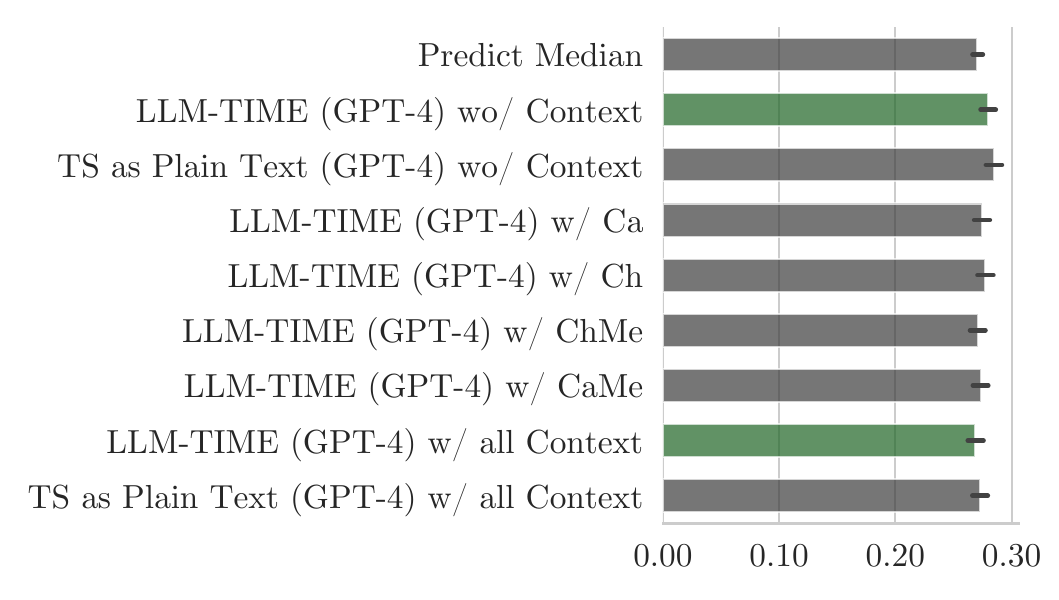}}
    \hfill
    \subfigure[MAE with z-score normalization]{\includegraphics[height=4cm,width=0.48\textwidth]{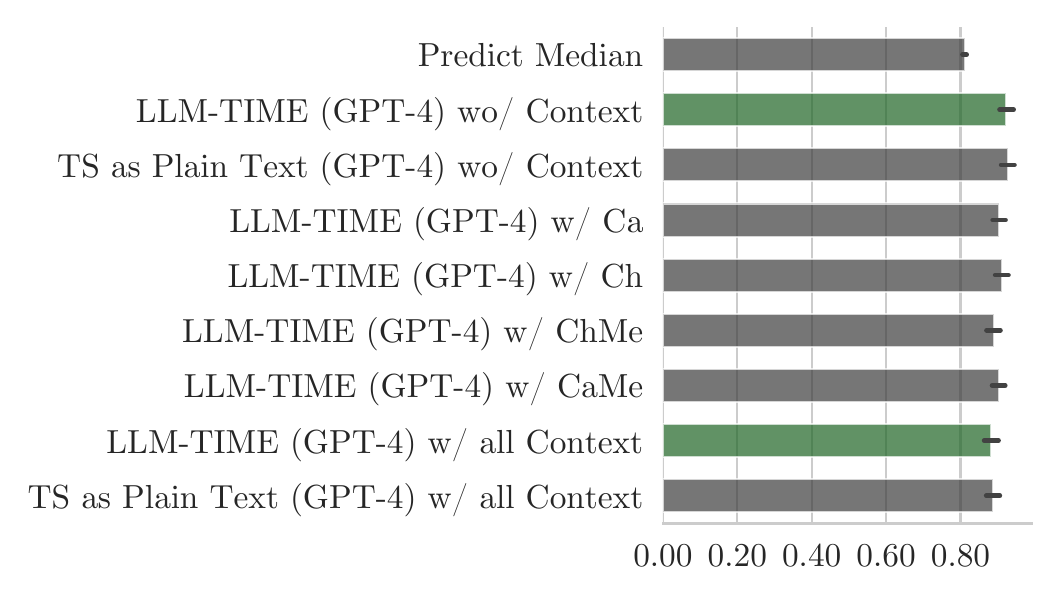}}
    \subfigure[MSE with max-min normalization]{\includegraphics[height=4cm,width=0.48\textwidth]{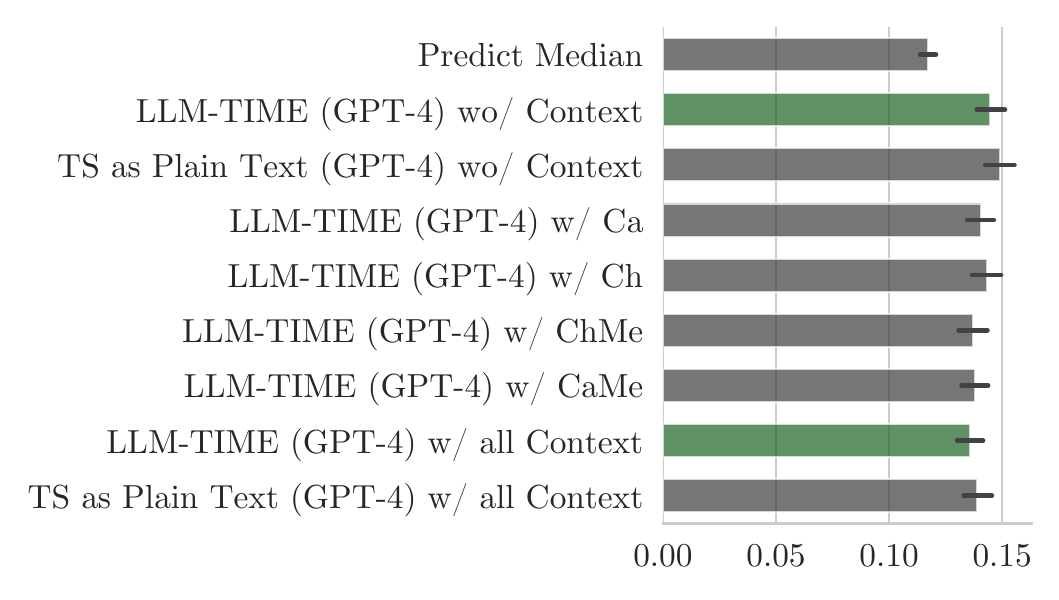}}
    \hfill
    \subfigure[MSE with z-score normalization]{\includegraphics[height=4cm,width=0.48\textwidth]{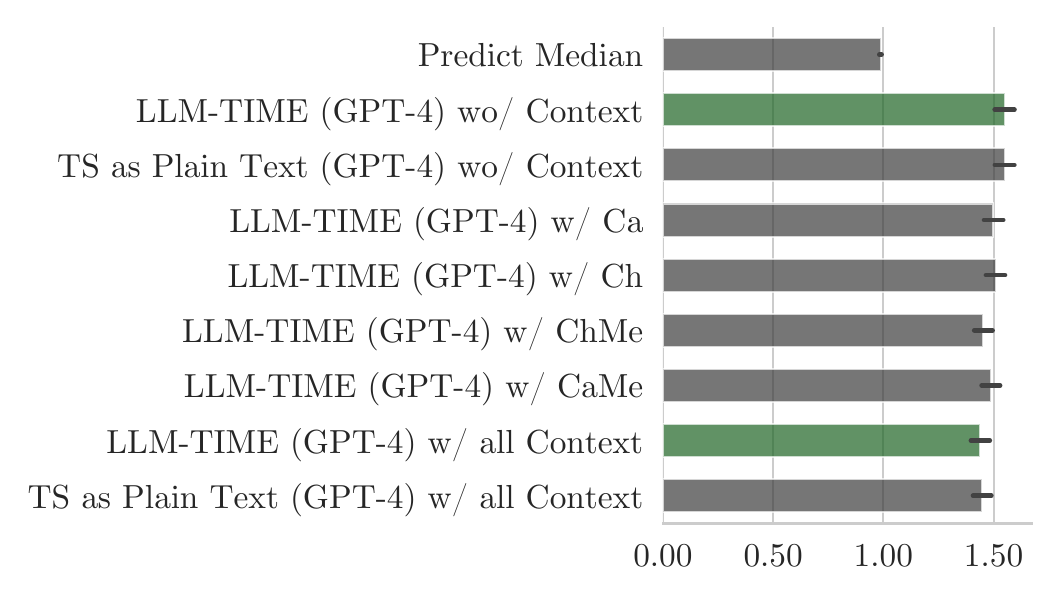}}
    \hfill
    \caption{These figures indicate that after adding various context relevant to the time series, the forecast results improved \textit{marginally}. We use "Predict Median", "\llmtime~\citep{gruver2023large} (GPT-4)", and "TS as Plain Text (GPT-4)" as our baselines. In the baseline, LLM forecasts without context (wo/ Context). It can be observed that whether providing Caption (Ca), Characteristics (Ch), or Metadata (Me) individually, such as "\llmtime (GPT-4) w/ Ca", or combining all captions, for example, "\llmtime (GPT-4) w/ all Context", the overall improvement remains very limited.}
     \label{fig:gpt4-forcast_full}
\end{figure}

\begin{figure}[t]
    \centering
    \subfigure[LLM reasoned out the distribution shift in the time series from the captions.]{\includegraphics[height=2.7cm,width=\textwidth]{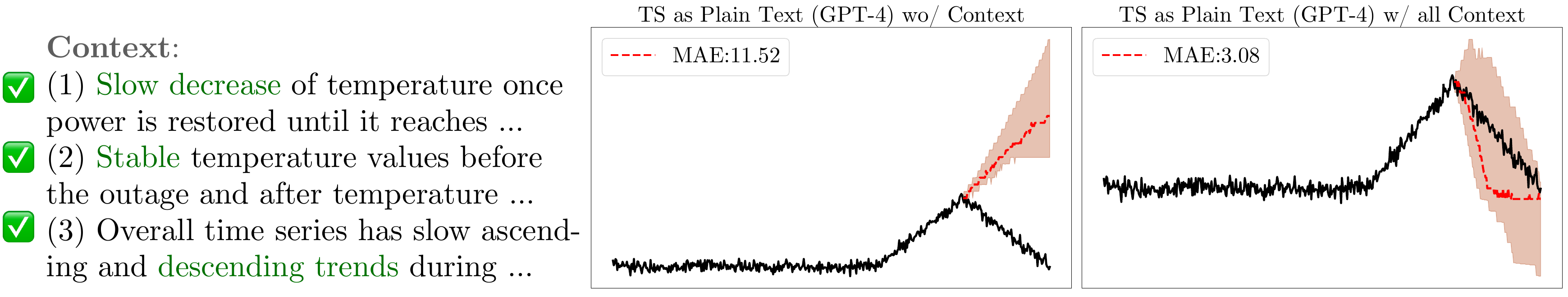}}
    \hfill
    \subfigure[Evne in a relatively simple pattern, the LLM fails to effectively understand captions.]{\includegraphics[height=2.7cm,width=\textwidth]{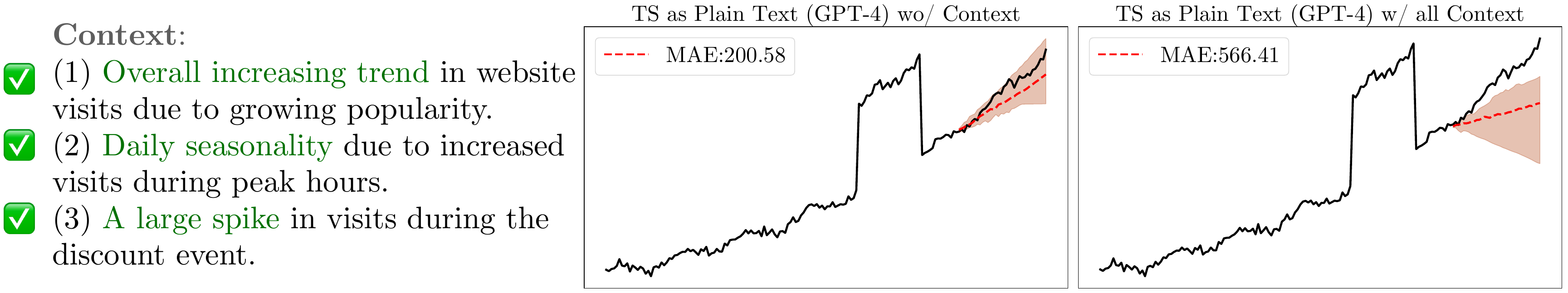}}
    \hfill
    \caption{Figures (a) and (b) are two typical examples showing that LLM can reason out difficult-to-forecast distribution shifts from captions. However, in a simple pattern, even when accurate captions are provided, it still fails to reason effectively.}
    \label{fig:helps_limited}
\end{figure}

\begin{figure*}
   \centering
    \includegraphics[width=0.98\linewidth]{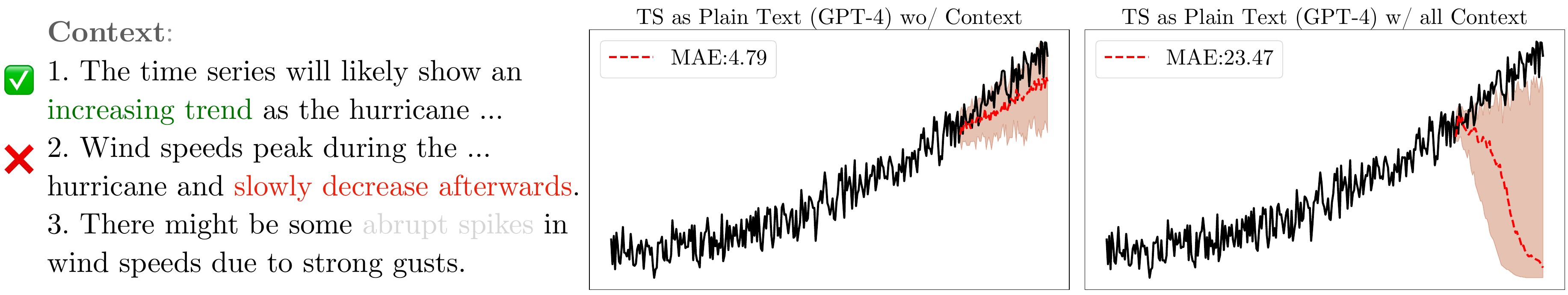}
    \caption{An example shows that LLMs, during the forecasting process, "misunderstood" the descriptions of time series trends in the captions, resulting in completely opposite reasoning.}
    \label{fig:text_help_hurt}
\end{figure*}

\begin{figure}[t]
    \centering
    \subfigure[For simple distribution shifting pattern, captions improves reasoning during forecasting.]{\includegraphics[height=2cm,width=\textwidth]{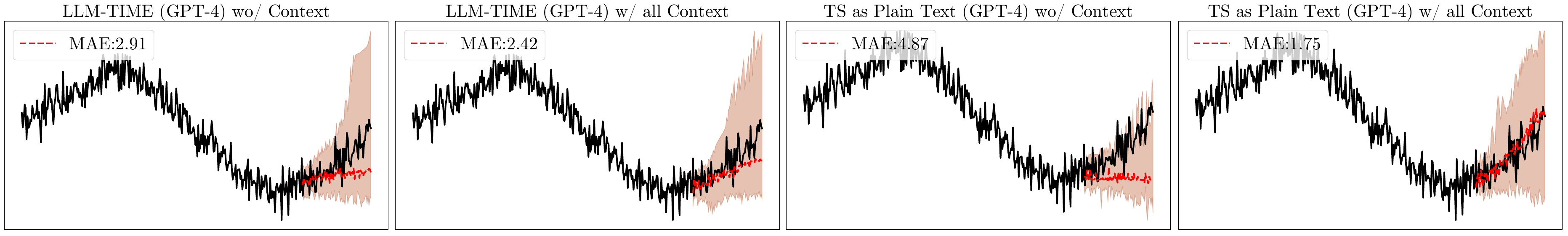}}
    \hfill
    \subfigure[For very difficult to forecast time series, captions still provide significant help to LLM reasoning.]{\includegraphics[height=2.1cm,width=\textwidth]{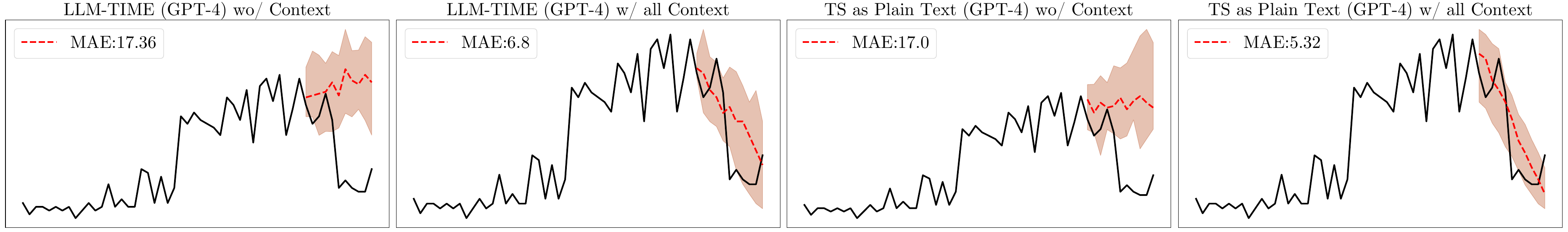}}
    \hfill
    \caption{Examples (a) and (b) show that integrating captions into forecasting, whether utilizing the \llmtime method or directly using \gptf, helps with LLM reasoning.}
    \label{fig:text_helps}
\end{figure}

In this section, we will present more results and examples on how LLM reasons through context in forecasting. \autoref{fig:gpt4-forcast_full} shows the full results for two metrics, MAE and MSE, both derived from the average of 2000 samples. Each result will be independently normalized before calculating the metrics. Overall, it can be seen that as more captions are provided, LLM's reasoning in forecasting only improves slightly. Even when all captions are provided, the aid remains quite \textit{marginal}. Two examples of how LLM integrates context into forecasting are shown in \autoref{fig:helps_limited}, where figure (a) demonstrates that LLM can reason out difficult-to-forecast distribution shifts from captions. However, as seen in figure (b), even when highly-relevant caption are provided, it still does not enhance the forecasting. There are even case like in \autoref{fig:text_help_hurt}, where LLM "misinterprets" the hints in the captions, leading to completely opposite conclusions. Additionally, even though current LLMs show quite limited zero-shot reasoning ability about time series, they still demonstrate \textit{somewhat potential}. Examples in \autoref{fig:text_helps} illustrate some successful cases. Therefore, we believe that with the development of general models, LLM's reasoning ability on numerical sequences, especially with natural language context, will gradually improve. 



\figAnnotatorTool
\figDonut